\documentclass[a4paper,conference]{IEEEtran}
\usepackage{amsmath,bm}
\usepackage{algorithm}
\usepackage{algorithmic}

\usepackage{booktabs}
\usepackage{caption}
\usepackage{multirow}

\usepackage{booktabs}
\usepackage{soul}
\usepackage{colortbl}

\usepackage{caption} 
\usepackage{graphicx}
\usepackage[hyphens]{url} 
\usepackage{cite}
\usepackage{hyperref}

\usepackage{colortbl}

\begin{document}

\title{GraSens: A Gabor Residual Anti-aliasing Sensing Framework for Action Recognition using WiFi}

\author{\IEEEauthorblockN{Yanling Hao}
\IEEEauthorblockA{ Queen Mary University\\
 of London\\
 London, UK\\
Email: yanling.hao@qmul.ac.uk}
\and
\IEEEauthorblockN{Zhiyuan Shi}
\IEEEauthorblockA{Onfido Research London\\
 London, UK\\
Email: zhiyuan.shi@onfido.com}
\and
\IEEEauthorblockN{Xidong Mu}
\IEEEauthorblockA{Queen Mary University\\
 of London\\
 London, UK\\
Email: x.mu@qmul.ac.uk}
\and
\IEEEauthorblockN{Yuanwei Liu}
\IEEEauthorblockA{ Queen Mary University\\
 of London\\
 London, UK\\
Email: yuanwei.liu@qmul.ac.uk}}

\maketitle

\begin{abstract}
WiFi-based human action recognition (HAR) has been regarded as a promising solution in applications such as smart living and remote monitoring due to the pervasive and unobtrusive nature of WiFi signals. However, the efficacy of WiFi signals is prone to be influenced by the change in the ambient environment and varies over different sub-carriers. To remedy this issue, we propose an end-to-end Gabor residual anti-aliasing sensing network (GraSens) to directly recognize the actions using the WiFi signals from the wireless devices in diverse scenarios. In particular, a new Gabor residual block is designed to address the impact of the changing surrounding environment with a focus on learning reliable and robust temporal-frequency representations of WiFi signals. In each block, the Gabor layer is integrated with the anti-aliasing layer in a residual manner to gain the shift-invariant features. Furthermore, fractal temporal and frequency self-attention are proposed in a joint effort to explicitly concentrate on the efficacy of WiFi signals and thus enhance the quality of output features scattered in different subcarriers. Experimental results throughout our wireless-vision action recognition dataset (WVAR) and three public datasets demonstrate that our proposed GraSens scheme outperforms state-of-the-art methods with respect to recognition accuracy.
\end{abstract}

\IEEEpeerreviewmaketitle

\section{Introduction}
Human action recognition (HAR) has attracted considerable attention in a range of applications, such as assisted living~\cite{wu2018wifi}, behavior analysis~\cite{wang2018csi}, and health monitoring~\cite{tan2018exploiting}. Many pioneering actions sensing attempts~\cite{lindell2019acoustic,isogawa2020optical,Li2019Making} have continuously emerged and developed in recent years to enhance measurement data and expand signal acquisition range~\cite{luo2021intelligent}. These sensing techniques motivate the breakthrough of long-time monitoring in a non-intrusive way~\cite{zhao2018through,wang2019can,wang2019person,Li2019Making}.

 The radio frequency (RF)-based technique is one of the most promising technologies among other action sensing technologies to localize people and track their motion~\cite{ li2016csi,qian2017inferring}. This attempt draws on the propagation of electromagnetic (EM) waves which are almost distributed at everyone's home. Benefit from the ubiquitous deployment, using WiFi signals for HAR in the indoor environment, is an economic solution~\cite{zhang2009dynamic,li2021wi}. Furthermore, WiFi-based solutions have no requirements of line-of-sight (LOS) thereby enabling larger detection areas than vision-based techniques~\cite{zhao2018through,wang2019can}. Therefore, WiFi-based HAR methods have received increasing attention~\cite{luo2021intelligent}. 

Extant researches have demonstrated the great potential of employing WiFi signals as a sensing approach~\cite{wang2019person}. Previously, most techniques for HAR are presented based on hand-crafted features from WiFi signals~\cite{adib2013see}. In essence, WiFi signals are susceptible to severe multipath and random noise in indoor surroundings. Hence, these manually designed features based mechanisms have certain limitations due to their heavy dependence on prior knowledge~\cite{li2016csi}. Furthermore, the efficacy of WiFi signals for HAR scatters over different sub-carriers since certain bands are sensitive to certain movements. Therefore, it is of vital importance to explore the problem of how to non-manually obtain robust and reliable representations from the WiFi signals. Deep learning is capable of automatic feature selection and has emerged as a new paradigm for mining the temporal-frequency information in the WiFi signals in diverse scenarios.
 
Deep learning has been evolving as a promising solution for HAR over the past few years~\cite{khong2018improving,sharif2014cnn}. Past deep learning methods however are prone to cause distortions after downsampling operation~\cite{taylor2010convolutional}. In deep learning networks, the downsampling operation is broadly utilized to reduce parameters and computation cost~\cite{pitas2000digital}. After the sampling operation, high-frequency information signals degenerate into completely different ones, which further disturbs the feature information~\cite{zou2020delving}. The standard solution of embedding a low-pass filter before sampling~\cite{zhang2019making} is unsatisfying because it degrades performance.
 
To remedy the above limitations, in this paper, an end-to-end Gabor residual anti-aliasing sensing (GraSens) network is proposed for HAR in varied environments. The architecture exploiting the reliable temporal-frequency representations from wireless signals is in an end-to-end style. The main contributions are summarized as follows:
\begin{figure*}[t]
    \centering
    \includegraphics[width=\textwidth]{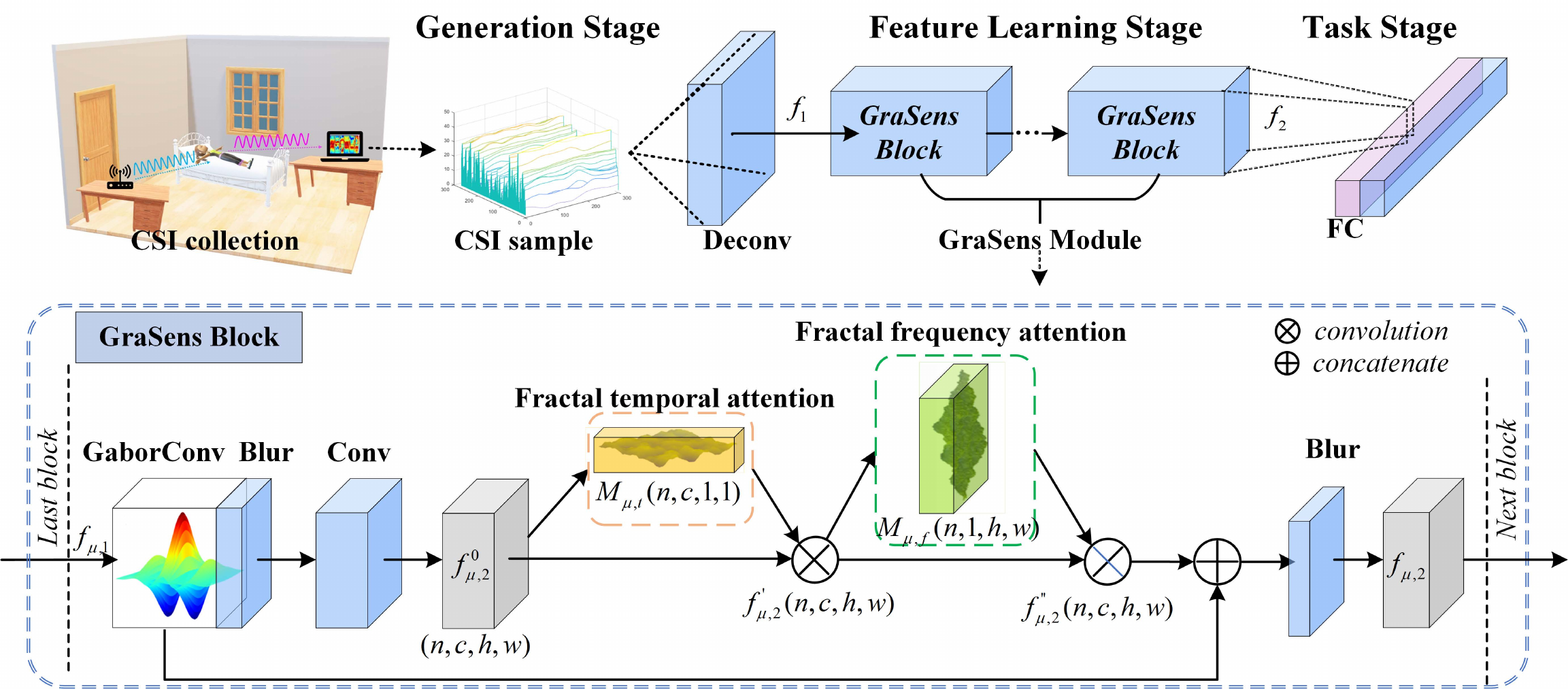}
    \caption{Overview of the proposed GraSens method.}
    \label{fig:architecture}
\end{figure*}

1) We propose a Gabor residual anti-aliasing sensing network to directly recognize the activities based on the WiFi signals from wireless devices such as smartphones and routers in diverse scenarios.

2) We design a Gabor residual block for exploiting reliable and robust WiFi signals representations to mitigate the influence of the change in the ambient environment. Specifically, the Gabor layer in this block is integrated with anti-aliasing operation in a residual manner to gain the shift-invariant features.

3) We design a fractal temporal and frequency self-attention mechanism to jointly explore the frequency and temporal continuity inside WiFi signals to enhance the quality of output features scattered in different subcarriers. 

4) We conduct experiments on our proposed wireless-vision action recognition dataset and the other three public datasets. The experimental results show that our method is robust over different scenes and outperforms competitive baselines with a good margin on the recognition accuracy.

\section{Related work}
Current researches on HAR can be loosely classified into two types, namely, video-based methods~\cite{pishchulin2016deepcut,isogawa2020optical} and RF-based methods~\cite{qian2017inferring}.
\subsection{Video-based human action recognition}
Video-based sensing methods have been prevailing in human action recognition. These methods capture image sequences by exploiting the camera and realize human action recognition using classification algorithms. Generally, they can be categorized into three groups: part-based frameworks~\cite{pishchulin2016deepcut}, two-step frameworks~\cite{isogawa2020optical}, multi-stream model frameworks. In the part-based HAR,  body parts are firstly detected separately and further assembled for human pose estimations such as DeepCut~\cite{pishchulin2016deepcut}. However, the assembled pose is prone to be ambiguous when more than one person gathers together and causes occlusion. Moreover, the part-based scheme is unable to recognize human pose globally since it focuses only on the second-order dependence of human body parts. As for the two-step framework, human bounding boxes are first detected and the poses within each box are then estimated such as Faster RNN~\cite{ren2015faster}. In this way, the quality of action recognition is highly attached to the accuracy of the detected human bounding boxes. In the presence of the multiple streams framework like RGB flow and optical flow,  it aims to improve the accuracy of action recognition by characterizing and integrating the patterns from various stream sources such as SlowFast~\cite{feichtenhofer2019slowfast}. However, most of the video-based methods are susceptible to ambient surroundings such as occlusion, lightning and privacy concerns, etc. To break the obstacles of the demand for line-of-sight (LOS),  a time-series generative adversarial network (TS-GAN)~\cite{gao2019know} is proposed to generate inferences and hallucinations in recognizing videos related to unseen actions. In fact, such hallucinations tend to produce errors due to the deformable ability of the human body. 

\subsection{WiFi based human action recognition}
RF-based techniques include radars~\cite{zhao2018through}, LiDARs~\cite{garcia2016pointnet} and WiFi devices~\cite{qian2017inferring}. Radar and LiDARs sensors demand dedicated and specially designed hardware. In contrast, WiFi devices are ubiquitously deployed since they are cost-effective and power-efficient. Besides, WiFi devices are free from the influences of illumination and privacy concerns in comparison to video-based methods. Recently, an amount of WiFi-based sensing systems were developed for human action recognition, such as WifiU~\cite{wang2016gait} and RT-Fall~\cite{wang2016rt}. Yet, previous systems are fairly coarse. These systems either locate only one single limb or produce a rough and static representation of the human body~\cite{qian2017inferring}. Most of the methods often target the general perception, for example, the rough classification~\cite{qian2017inferring} and indoor localization~\cite{adib2013see}. To mitigate the situation, some researchers attempt to simulate 2D or 3D skeletons based on wireless signals for person perception~\cite{luo2021intelligent}. Other researchers simulate the WiFi arrays to enhance the accuracy of recognition and localization~\cite{holl2017holography}. These researches illuminate the optimizing applications of WiFi-based HAR in varied environmental conditions. Recently, Alazrai et al. proposed an end-to-end framework E2EDLF~\cite{alazrai2020end} to recognize human-to-human interactions by sophisticated and careful construction of the input CSI image.

\section{Architecture for WiFi Sensing via Gabor Residual Anti-Aliasing}
As seen in Fig.~\ref{fig:architecture}, the proposed GraSens is designed and conceived to fully exploit and explore the data collected from off-the-shelf commercial WiFi devices in an end-to-end style. Three stages can be generalized, namely generation stage, feature learning stage, and task stage. Specifically, the generation stage is aiming to enable the raw WiFi channel state information (CSI) data compatible with the input of the network while preserving the original frequency and temporal information. The feature learning stage as shown in the bottom part of Fig.~\ref{fig:architecture} is defined as Gabor residual anti-aliasing attention module, which puts forward the up-sampled CSI samples for feature maps generation. This stage can greatly mitigate the influence of the ambient noises that are confused with the action signals, and improve the quality of output features from CSI information scattered in different subcarriers. These learned features are further fed to fully connected layers for a particular task in the last stage.

\subsection{The proposed GraSens network}
\label{sec:mainpipeline}
\subsubsection{Generation Stage}
\begin{figure}[t]
    \centering
    \includegraphics[scale=0.14]{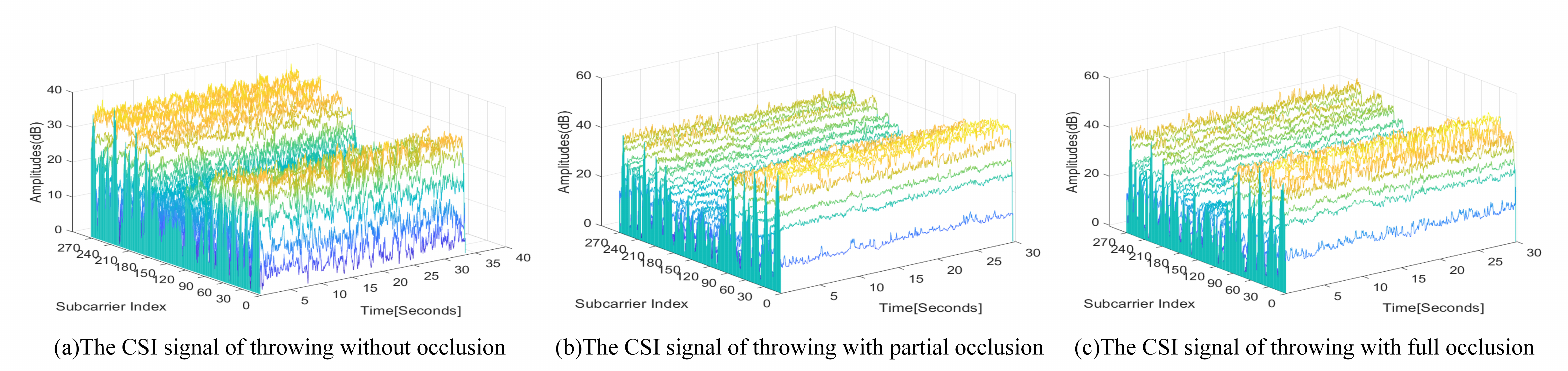}
    \caption{The CSI signal of throwing. (a)-(c) are the CSI signals of throwing in scenes without occlusion, with partial occlusion, and with full occlusion, respectively.}
    \label{fig:csi_pre}
\end{figure}

To preserve the temporal as well as frequency information within the CSI signals, the raw CSI signals are transformed into a set of CSI tensors with learnable parameters in the generation stage seen in Fig.~\ref{fig:csi_pre}(a). 
Firstly, the raw CSI signals of an action segment as shown in Fig.~\ref{fig:architecture} are converted into a series of CSI tensors, aiming to interpret the action with multiple aspects. After this, all the CSI tensors are up-sampled by the deconvolution operation adapted to the network. The principle of WiFi-based sensing is to recognize the influence of perceived objects on the transmitted signals~\cite{wang2016rt}. Generally, a WiFi system can be modeled and summarised as follows:
\begin{equation}
{\bm{B}_s}(i) = {\bm{\gamma}_s}(i){\bm{A}_s}(i) + \bm{\theta},
\end{equation}
where $s\in[1,\cdots,N_s]$ depicts the index of the orthogonal frequency-division multiplexing (OFDM) subcarriers employed in the WiFi device, ${N_s}$ represents the total number of the OFDM subcarriers. $i$ defines the index of the transmitted and received packets. The ${i^{th}}$ transmitted and received packets pertinent to the OFDM subcarrier frequency $s$ are specified as ${\bm{A}_s}(i)$ and ${\bm{B}_s}(i)$, respectively. $\bm{\theta}$ represents the received noise, and a complex-valued matrix ${\bm{\gamma}_s}$ constitutes the CSI measurements for the OFDM subcarrier frequency $s$. ${\bm{\gamma}_s}$ is of dimensions ${N_T} \times {N_R}$ whose horizontal and vertical coordinates indicate the number of transmitting and receiving antennas, respectively. 

In each time serial sequence, the raw CSI signals are arranged in a 2D matrix of dimensions $\bm{\gamma} \times I$ with $\bm{\gamma} = {N_T} \times {N_R} \times {N_S}$ where $I$ indicates the index of packets recorded in a specific CSI time serial. A sliding window along the time axis divides the CSI signals into a bank of overlapped segments as CSI tensor $CSI(i)$ of size $\phi\times \bm{\gamma} $. $\phi $ defines the number of packets and $\upsilon $ implies the overlap between every two adjacent segments, where $\upsilon \le \phi $ and $i\le I/ \upsilon$. 

 The CSI samples are further put forward to the deconvolution layer. The deconvolution layer serves as an up-sampling layer to up-sample feature maps of the input CSI tensor and preserves the connectivity pattern. In the up-sampling process, the input CSI tensor is enlarged and densified by cross-channel convolutions with multiple filters. The spatial and frequency information in each channel is expanded and encoded into spatially-coded maps. In comparison with the extant resizing methods, the benefit of the deconvolution layers is that the parameters are trainable. During the training, the weights of deconvolution layers are constantly updated and refined. The CSI samples are up-sampled to be processed by feature learning modules as follows:
\begin{equation}
\label{eq:2}
{f_1} = Deconv(CSI).
\end{equation}
where $Deconv(\cdot)$ is the deconvolution operation.


\subsubsection{Feature Learning Stage}
As depicted in Fig.~\ref{fig:architecture}, a Gabor residual anti-aliasing sensing module is proposed for shift-invariant feature learning. This GraSens module consists of several Gabor residual anti-aliasing blocks. In each block, a Gabor convolution layer filter replaces the first convolution layer in a traditional residual module and serves as initialization to gain more discriminative power. After this, an anti-aliasing layer is further added to keep the output feature maps shift-invariant. For block $\mu$, given the intermediate feature map ${f_1}  \in {\mathcal{R}^{C \times H \times W}}$ as the input, the output features can be generated as follows:
\begin{equation}
\label{eq:3}
{f_{\mu,2}^{0}} = Conv(Blur(GaborConv({f_1}))).
\end{equation}
where $GaborConv(\cdot)$ is the Gabor convolution operation and $Blur(\cdot)$ is the anti-aliasing operation.
To explicitly concentrate on the efficacy of WiFi signals, GraSens sequentially infers a 1D fractal dimension based temporal attention map $M_{\mu ,t} \in {R^{C \times 1 \times 1}}$ and a 2D fractal dimension based frequency attention map $M_{\mu ,f} \in {R^{C \times H \times W}}$ as shown in Fig.~\ref{fig:architecture}. In short, the whole attention process can be generalized as follows:
\begin{equation}
\label{eq:4}
\begin{array}{l}
f_{\mu ,2}^{'} = {M_{\mu ,t}}({f_{\mu ,2}}) \otimes {f_{\mu ,2}},\\
f_{\mu ,2}^{''} = {M_{\mu ,f}}(f_{\mu ,2}^{'}) \otimes f_{\mu ,2}^{'},
\end{array}
\end{equation}
where $\otimes$ indicates the element-wise multiplication. The unique asset of multiplication locates in the way of duplication of attention values. Intuitively, temporal attention values replicated along the frequency axis and vice versa. Herein, the refined output $f_{\mu ,2}$ of stacked block $\mu$ can be formulated as follows:
\begin{equation}
\label{eq:5}
f_{\mu ,2} = Blur(f_{\mu ,2}^{''} \oplus f_1),
\end{equation}
where $\oplus$ is the concatenate operation. Fig.~\ref{fig:architecture} describes the calculation process of each attention map. After several blocks, ${f_{2}}$ is the final output temporal and frequency representation. The following section~\ref{sec:GraSens} describes the details of each attention module. The feature learning progress of GraSens module is as depicted in Algorithm~\ref{learning}.
\begin{algorithm}
    \caption{Feature Learning}
    \label{learning}
    \hspace*{\algorithmicindent} \textbf{Input:} The up-sampled CSI sample ${f_1}$  \\
    \hspace*{\algorithmicindent} \textbf{Output:} The output feature maps ${f_{2}}$ of GraSens module
    
    \begin{algorithmic}[1]
    \STATE Choose the number of stacked GraSens blocks as $\lambda$;
    \STATE Initialize the block $\mu =1$;
    \REPEAT   
    \FOR{block $\mu$}
    \STATE Update the Gabor anti-aliasing output ${f_{\mu,2}^{0}} \gets {f_1}$ using Eqs.~\eqref{eq:3},~\eqref{eq:8} and~\eqref{eq:9};
    \STATE Update the fractal self-attention output $f_{\mu ,2}^{''}\gets {f_{\mu,2}^{0}}$ using Eqs.~\eqref{eq:10}-~\eqref{eq:12};
    \STATE Update the anti-aliasing output $f_{\mu ,2} \gets f_{\mu ,2}^{''}$ using Eqs.~\eqref{eq:5} and~\eqref{eq:9};
    \ENDFOR
    \STATE Renew the input for next block ${f_1} = f_{\mu ,2}$;
    \STATE Move to next block $\mu = \mu +1$;
    \UNTIL $\mu = \lambda$;
    \STATE Return ${f_{2}} = f_{\lambda ,2}$ and forward to the task stage.
    \end{algorithmic}
\end{algorithm}
\subsubsection{Task Stage}
During the task stage, the learned frequency and temporal features are fed to one fully connected layer to generate the outputs for a particular task. In the training of GraSens, the loss is computed by the activation function and loss function. In this way, the difference between the outputs of the GraSens network ${f_3} $ and the ground-truth G can be measured by the loss. The output ${f_3} $ is formulated as follows:
\begin{equation}
{f_3} = Blur(FC({f_2})),
\end{equation}
The cross-entropy loss is a basic option to be applied to optimize GraSens and given by:
\begin{equation}
   \mathcal{L} =\sum\limits_{j = 1}^J {{{f_3}^j}\log ({G^j})}.
\end{equation}
where $j$ is the snippet number of input training CSI samples. In addition, we utilize the Stochastic Gradient Descent with Momentum to learn the parameters.

\begin{table}[!ht]
\caption{Classification accuracy of the dataset WVAR.}
\label{tab:WVAR}
\tiny
\centering
\begin{tabular}{@{}lcccccccccc@{}}
\toprule
\textbf{Methods} & \textbf{\begin{tabular}[c]{@{}c@{}}fall\_\\down\end{tabular}} & \textbf{throw} & \textbf{push} & \textbf{kick} & \textbf{punch} & \textbf{jump} & \textbf{\begin{tabular}[c]{@{}c@{}}phone\_\\ talk\end{tabular}} & \textbf{seat} & \textbf{drink} & \textbf{OA}   \\ \midrule
\textbf{SVM}     & \textbf{1.00}         & 0.92           & 0.90           & 0.94          & 0.93           & 0.94          & 0.91                 & 0.88 & \textbf{1.00}     & 0.94          \\
\textbf{WNN}  & \textbf{1.00}         & \textbf{1.00}     & \textbf{1.00}    & 0.86          & 0.88           & \textbf{1.00}    & \textbf{1.00}           & 0.81          & \textbf{1.00}     & 0.94          \\
\textbf{GraSens}  & \textbf{1.00}               & \textbf{1.00}           & 0.95    & \textbf{0.97} & \textbf{0.99}     & \textbf{1.00}          & 0.88                  & \textbf{0.90}          & 0.92           & \textbf{0.95} \\ \bottomrule
\end{tabular}
\end{table}
\begin{table}[h!]
\caption{Classification accuracy of the dataset WAR.}
\label{tab:WAR}
\centering
\scriptsize
\begin{tabular}{@{}lccccccc@{}}
\toprule
\textbf{Methods}     & \textbf{lie\_down} & \textbf{fall} & \textbf{run}  & \textbf{sit\_down} & \textbf{stand\_up} & \textbf{walk} & \textbf{OA}   \\ \midrule
\textbf{RF}~\cite{ho1995random}   & 0.53              & 0.60          & 0.81          & 0.88              & 0.49              & 0.57          & 0.65          \\
\textbf{HMM}~\cite{eddy2004hidden}  & 0.52              & 0.72          & 0.92          & 0.96              & 0.76              & 0.52          & 0.73          \\
\textbf{LSTM}~\cite{yousefi2017survey} & \textbf{0.95}              & 0.94          & \textbf{0.97} & 0.81              & 0.83              & \textbf{0.93} & 0.91          \\
\textbf{SVM}         & 0.91              & 0.96          & 0.93          & 0.96              & 0.71              & 0.87          & 0.93          \\
\textbf{WNN}      & 0.93              & 0.93          & 0.93          & \textbf{0.98}     & 0.90              & 0.86          & 0.95          \\
\textbf{GraSens}      & 0.94     & \textbf{0.97} & 0.95 & \textbf{0.98}              & \textbf{0.91}     & 0.85 & \textbf{0.96} \\ \bottomrule
\end{tabular}
\end{table}

\subsection{GraSens Module}
\label{sec:GraSens}
\subsubsection{Gabor Filtering based Anti-aliasing}
As for each GraSens block, the Gabor layer builds a convolution kernel library for feature extraction. To obtain the strong auxiliary feature information, the Gabor convolution kernel group is optimized by the network training and further convolved with the CSI samples. Generally, the Gabor function describes a complex sinusoid modulated by Gaussian in accordance with monotonicity and differentiability, i.e.,
\begin{equation}
\label{eq:8}
 \begin{array}{l}
 {GaborConv}=g(x,y,\varpi ,\theta ,\psi ,\sigma ) \\ =  \exp ( - \frac{{x{'^2} + y{'^2}}}{{2{\sigma ^2}}})\cos (\varpi x' + \psi ),
 \end{array}
\end{equation}
where $ x' = x\cos \theta  + y\sin \theta$, and $y' =  - x\cos \theta  + y\cos \theta$. Gabor layers prove to be efficient for spatially localized features extracting~\cite{luan2018gabor}. To extract the features from the WiFi signals, a set of Gabor filters are used as ref~\cite{alekseev2019gabornet}. Frequencies ${\varpi _n}$ of the Gabor filters is obtained by $ {\varpi _n} = \frac{\pi }{2}{\sqrt 2 ^{ - (n - 1)}}$ , $n = 1,2, \ldots ,5$. The orientations ${\theta _m}$ is set as ${\theta _m} = \frac{\pi }{8}(m - 1)$ , where $ m = 1,2, \ldots ,8$. In addition, the ${\sigma} $ is defined by the relationship between $\sigma $ and $\varpi $ where $\sigma  \approx \frac{\pi }{\varpi }$. $\psi $ follows the uniform distribution ${\rm{U(0, }}\pi {\rm{)}}$. Accordingly, the Gabor Layer weights in this paper are initialized similarly.

Subsequently, the anti-aliasing layer is leveraged to enable the extracted feature shift-invariant. The anti-aliasing layer serves as two steps. To begin with, a set of low-pass filters $\Psi $ are arranged and generated in terms of varied spatial locations and channel groups within each GraSens block. After than, the predicted filters are adopted and applied back onto the input feature maps on account of anti-aliasing. We assume an input feature $X$. To be specific, a low-pass filter ${\Psi _{i,j}^{p,q}}$, for example, a 3×3 convolution filter, is generated to down-sample the input feature $ X $ over each spatial location $(i, j)$ as follows: 
\begin{equation}
\label{eq:9}
Blur = \sum\limits_{p,q \in \Omega } {\Psi _{i,j}^{p,q} \cdot {X_{i + p,j + q}}}.
\end{equation}

\subsubsection{Fractal Dimension based Self-Attention}
Fractal describes unusual objects of irregular shapes which have a high degree of complex properties. Fractal dimension can indicate the degree of the complexity of objects, such as the irregular WiFi signals. For the convenience, a general expression has been defined to measure the fractal dimension as follows:
\begin{equation}
\label{eq:10}
   FD =-\mathop {lim}\limits_{\varepsilon  \to 0}\frac{{log(\eta (\varepsilon ))}}{{log(\varepsilon )}},
\end{equation}
where $\eta$ measures self-similarity and $\varepsilon $ denotes the scale. In our work, $FD$ is employed to calculate the fractal dimension of feature maps along with the frequency and temporal domain. 

\noindent \textbf{\textit{Fractal temporal attention module.}}
Each channel within a feature map can reflect the diverse temporal characteristics of the input CSI samples. Inspired by the CBAM~\cite{woo2018cbam}, we calculate the fractal dimensions for all the frequencies in feature maps input as the temporal attention as follows:
\begin{equation}
\label{eq:11}
{M_{\mu ,t}}({f_{\mu ,2}}){\rm{ }} = \xi (MLP(FD({f_{\mu ,2}}))),
\end{equation}
where $\xi$ implies the sigmoid function. $MLP$ specifies a multi-layer perceptron operation. 

\noindent \textbf{\textit{Fractal frequency attention module.}}
Cross-channels within a feature map can capture the frequency characteristics. For this purpose, a frequency attention map is generated to exploit the cross-channel relationship of features. Fractal dimensions across the channel are utilized to generate one feature map as the fractal feature maps. Those fractal feature maps are further fed to a standard convolution layer and thus generate the frequency attention map. In brief, the fractal frequency attention is calculated as follows:
\begin{equation}
\label{eq:12}
{M_{\mu ,f}}(f_{\mu ,2}^{'}){\rm{ }} = \xi ({Conv}(FD(f_{\mu ,2}^{'}))),
\end{equation}
where $Conv$ represents a convolution operation.

\begin{figure}[t]
    \centering
    \includegraphics[scale=0.38]{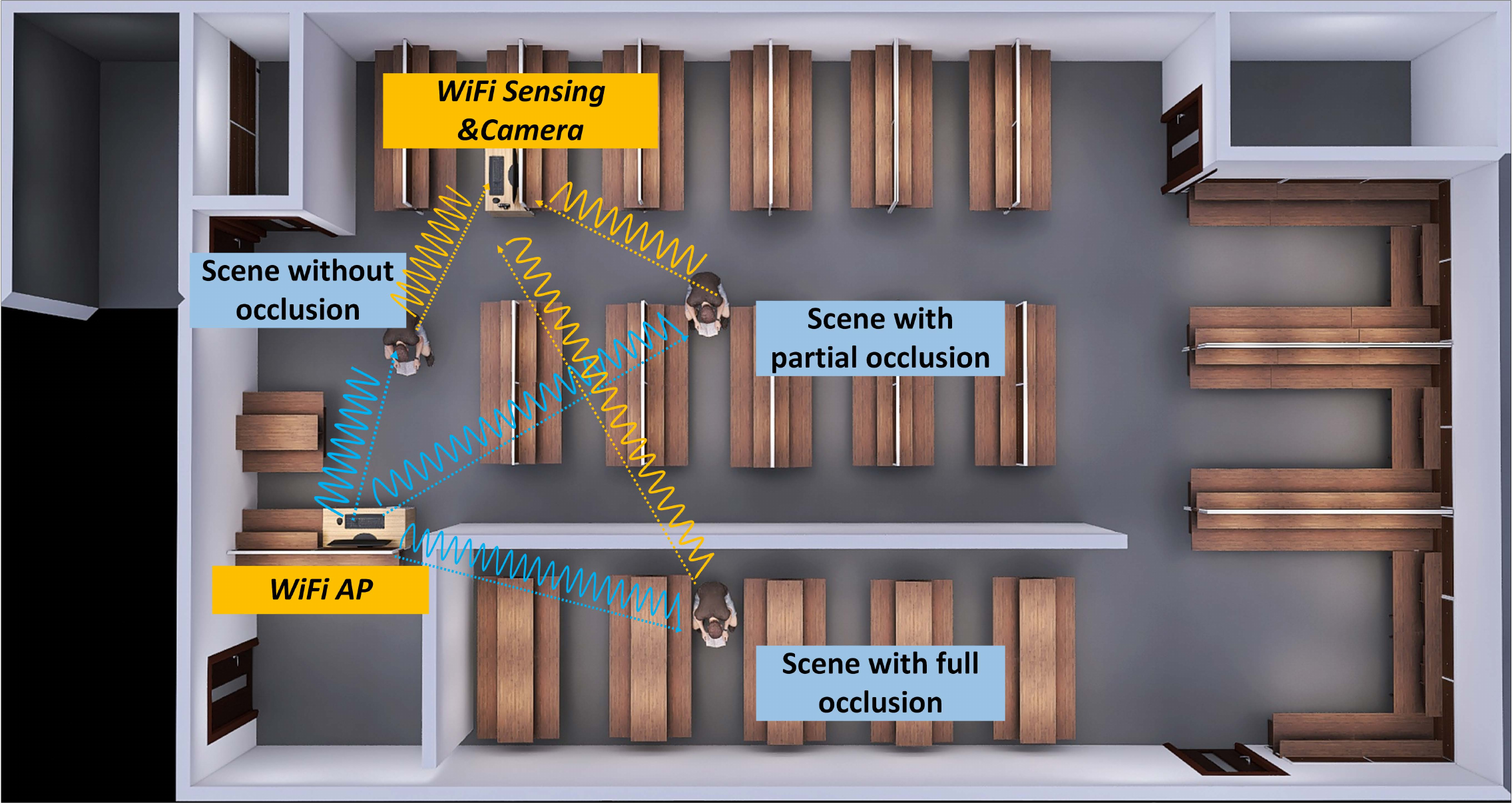}
    \caption{Three experiment scenes of WVAR dataset.}
    \label{fig:scene}
\end{figure}

\section{Experiments}
\subsection{Datasets}
\begin{table*}[h!]
\caption{Classification accuracy of the dataset HHI.}
\label{tab:HHI}
\tiny
\centering
\resizebox{\textwidth}{!}{
\begin{tabular}{@{}lccccccccccccc@{}}
\toprule
\textbf{Methods}           & \textbf{approaching} & \textbf{departing} & \textbf{\begin{tabular}[c]{@{}c@{}}hand\_\\ shaking\end{tabular}} & \textbf{high five} & \textbf{hugging} & \textbf{\begin{tabular}[c]{@{}c@{}}kicking\_\\ left\_leg\end{tabular}} & \textbf{\begin{tabular}[c]{@{}c@{}}kicking \_\\ right\_leg\end{tabular}} & \textbf{\begin{tabular}[c]{@{}c@{}}pointing\_\\ left\_hand\end{tabular}} & \textbf{\begin{tabular}[c]{@{}c@{}}pointing\_ \\ right\_hand\end{tabular}} & \textbf{\begin{tabular}[c]{@{}c@{}}punching\_ \\ left\_hand\end{tabular}} & \textbf{\begin{tabular}[c]{@{}c@{}}punching\_\\  right\_hand\end{tabular}} & \textbf{pushing} & \textbf{OA}   \\ \midrule
\textbf{GoogleNet}~\cite{szegedy2015going}   & 0.93                 & 0.93               & 0.79                 & 0.76               & 0.64             & 0.54                                                                   & 0.50                                                                     & 0.78                                                                     & 0.77                                                                       & 0.59                                                                      & 0.59                                                                       & 0.68             & 0.71          \\
\textbf{ResNet-18}~\cite{he2016deep}   & 0.92                 & 0.90               & 0.85                 & 0.79               & 0.77             & 0.68                                                                   & 0.60                                                                     & 0.82                                                                     & 0.80                                                                       & 0.60                                                                      & 0.65                                                                       & 0.76             & 0.76          \\
\textbf{Squeeze-Net}~\cite{iandola2016squeezenet}   & 0.95                 & 0.93               & 0.83                 & 0.76               & 0.70             & 0.66                                                                   & 0.62                                                                     & 0.78                                                                     & 0.79                                                                       & 0.60                                                                      & 0.72                                                                       & 0.74             & 0.76          \\
\textbf{E2EDLF}~\cite{alazrai2020end}     & 0.96          & 0.92          & 0.89          & 0.84          & 0.86          & \textbf{0.78}      & \textbf{0.82}        & \textbf{0.85}        & \textbf{0.90}          & 0.73                  & \textbf{0.80}          & 0.86          & 0.85          \\
\textbf{SVM}               & \textbf{0.99}        & 0.96               & 0.90        & 0.83               & 0.82             & 0.73                                                                   & 0.79                                                                     & 0.69                                                                     & 0.62                                                                       & \textbf{0.74}                                                             & 0.77                                                                       & 0.74             & 0.78          \\
\textbf{WNN}            & 0.97                 & 0.96               & 0.83                 & 0.84               & 0.72             & 0.52                                                                   & 0.65                                                                     & 0.76                                                                     & 0.81                                                                       & 0.63                                                                      & 0.69                                                                       & 0.78             & 0.79          \\
\textbf{GraSens}             & \textbf{0.99}          & \textbf{0.97} & \textbf{0.91} & \textbf{0.89} & \textbf{0.89} & 0.58               & 0.68                 & 0.83                 & 0.79                   & 0.55                  & 0.75                   & \textbf{0.93} & \textbf{0.86}         \\ \bottomrule
\end{tabular}}
\end{table*}
\begin{table}[h!]
\centering
\caption{Classification accuracy of the dataset CSLOS}
\label{tab:LOS}
\tiny
\begin{tabular}{l!{\vrule width \lightrulewidth}l!{\vrule width \lightrulewidth}ccccccc} 
\toprule
\textbf{Scenes}              & \textbf{Methods}  &\textbf{\begin{tabular}[c]{@{}c@{}}no\_\\ move\end{tabular}} & \textbf{falling} & \textbf{walking} &\textbf{\begin{tabular}[c]{@{}c@{}}sitting/\\ standing\end{tabular}}  & \textbf{turning} & \textbf{\begin{tabular}[c]{@{}c@{}}picking\_\\up\end{tabular}} & \textbf{Average}  \\ 
\midrule
\multirow{4}{*}{\textbf{E1} }    & \textbf{SVM}~\cite{baha2021exploiting} & 0.98          & 0.86          & \textbf{1.00}          & 0.91          & 0.90          & 0.92          & 0.94             \\
                & \textbf{WNN}    & 0.89          & 0.80          & 0.73          & 0.86          & 0.67          & 0.94          & 0.81             \\
                & \textbf{GraSens}    & \textbf{0.97} & \textbf{0.97} & 0.95 & \textbf{0.98} & \textbf{0.96} & \textbf{0.99} & \textbf{0.97}                 \\
                             &                   &             &             &             &             &             &             &                   \\
\multirow{4}{*}{\textbf{E2}}     & \textbf{SVM}~\cite{baha2021exploiting} & \textbf{0.95} & 0.82          & \textbf{0.99} & 0.82          & 0.81          & 0.82          & 0.89             \\
                & \textbf{WNN}    & 0.84          & 0.78          & 0.75          & 0.83          & 0.69          & 0.84          & 0.79             \\
                & \textbf{GraSens}    & 0.93          & \textbf{0.94} & 0.98          & \textbf{0.91} & \textbf{0.92} & \textbf{0.91} & \textbf{0.93}                     \\
\bottomrule
\end{tabular}
\end{table}

\noindent\textbf{Our WVAR dataset}.
WVAR collection was implemented in one spacious office apartment by 2 volunteers who performed 9 activities with five repeated trials in different simulating occlusion occasions as seen in Fig.~\ref{fig:scene}. The experimental hardware as seen in Fig.~\ref{fig:architecture} constitutes two desktop computers as transmitter and receiver, both of which are carried out in IEEE 802.11n monitor mode operating at 5.4 GHz with a sampling rate of 100 Hz. WVAR also contains the synchronized video data recorded at 20 FPS, i.e. every frame is corresponding to five CSI packets.

Table~\ref{tab:LOS} shows the classification accuracy of the dataset CSNLOS. We test two LOS scenarios' data E1 and E2. The results of GraSens rank first compared to all other two methods in two LOS scenes E1 and E2. As for E1, GraSens achieves the best results by 3$\%$ average accuracy higher than SVM~\cite{baha2021exploiting}. With regard to E2, the performance of GraSens is better except for no\_movement and walking which still are comparable with those of SVM~\cite{baha2021exploiting}. In other words, GraSens has good robustness in comparison to the other two models.

\noindent\textbf{WAR, HHI, and CSLOS}.
The public available dataset WAR~\cite{yousefi2017survey} consists of 6 persons, 6 activities with 20 trials for each in an indoor office. The sampling rate is 1 kHz. 

The publicly available CSI dataset of HHIs~\cite{alazrai2020dataset} is composed of 12 different human-to-human interactions (HHI) which performed by 40 distinct pairs of subjects in an indoor environment inside an office with 10 different trials, e.g. approaching, departing, hand\_shaking, etc. 

Another public available cross-scene dataset (CSLOS)~\cite{baha2020dataset} is provided by the same group as the HHI. LOS contains five experiments in three different indoor environments, where two are of LOS nature and the third environment is of a non-line-of-sight (NLOS) nature. 30 different subjects were included with 20 repeated trials for each of the experiments in terms of the variations of human movements.


\subsubsection{Evaluation Metrics}
Accuracy and precision are utilized in the sort of performance evaluation. Accuracy defines the percentage of total actions classified correctly. Precision reflects the correct percentage of classified actions from all predicted ones. It should be underlined that false positives are also included in precision. Both metrics are denoted as follows:
$Accuracy =\frac{{TP + TN}}{{TP + TN + FP + FN}}$ and $Precision = \frac{{TP}}{{TP + FP}}$, where $TP$, $FP$, $TN$ and $FN$ represent the true and false positives and negatives, respectively.
\begin{figure*}[h]
    
    \centering
    \includegraphics[width=\textwidth]{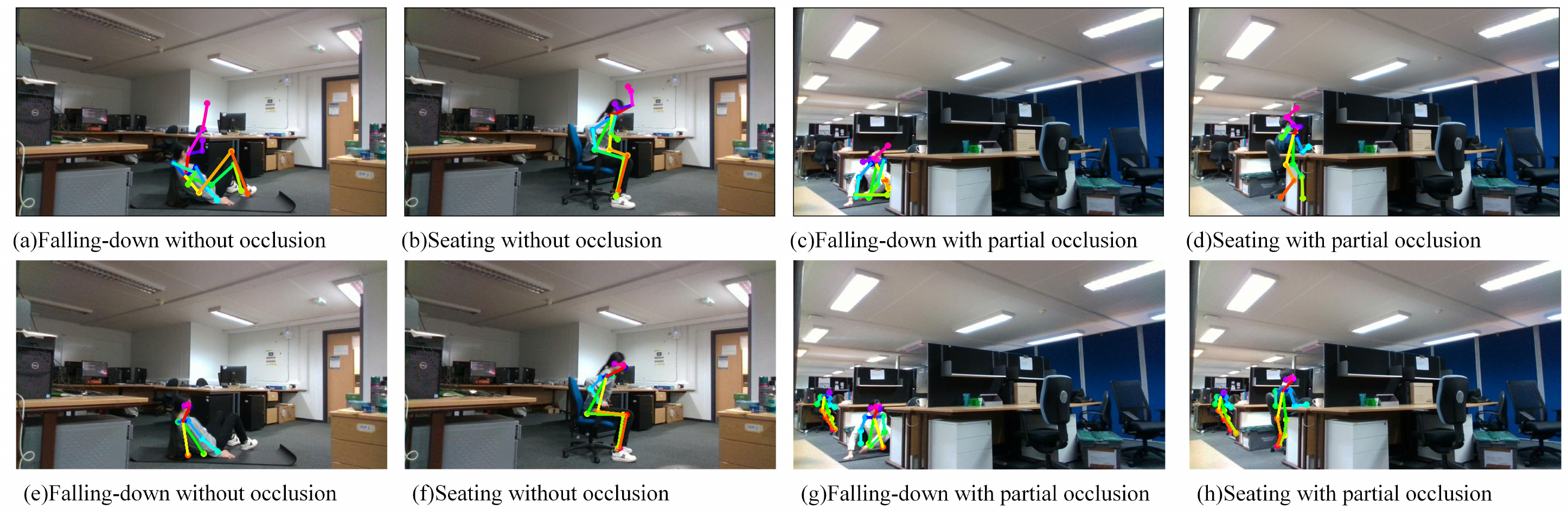}
    \caption{The skeleton results by WiFi (a)-(d)  and by video (e)-(h). In the scene without occlusion as the first two columns show, the skeleton results by WiFi are comparable in seating, and better in self-occlusion cases like falling down than those by video. As for the scene without occlusion in the last two columns, the skeleton results by WiFi are more precise seen in the legs in (d) compared to (h) and have less false detection like the chairs than those by video.}
    \label{fig:skeleton}
\end{figure*}
\begin{table*}[]
\caption{Ablation study of the number of GraSens blocks}
\label{tab:blocks}
\resizebox{\textwidth}{!}{
\begin{tabular}{@{}llllllllllllll@{}}
\toprule
\textbf{Blocks}    & \textbf{approaching} & \textbf{departing} & \textbf{\begin{tabular}[c]{@{}l@{}}hand\_\\ shaking\end{tabular}} & \textbf{high five} & \textbf{hugging} & \textbf{\begin{tabular}[c]{@{}l@{}}kicking\_\\ left\_leg\end{tabular}} & \textbf{\begin{tabular}[c]{@{}l@{}}kicking \_\\ right\_leg\end{tabular}} & \textbf{\begin{tabular}[c]{@{}l@{}}pointing\_\\ left\_hand\end{tabular}} & \textbf{\begin{tabular}[c]{@{}l@{}}pointing\_\\ right\_hand\end{tabular}} & \textbf{\begin{tabular}[c]{@{}l@{}}punching\_\\ left\_hand\end{tabular}} & \textbf{\begin{tabular}[c]{@{}l@{}}punching\_\\ right\_hand\end{tabular}} & \textbf{pushing} & \textbf{OA}   \\ \midrule
\textbf{$\lambda = 4$}  & 0.96                 & \textbf{0.98}      & 0.84                                                              & 0.80                & 0.70              & 0.50                                                                    & 0.49                                                                     & 0.83                                                                     & \textbf{0.84}                                                             & \textbf{0.65}                                                            & \textbf{0.81}                                                             & \textbf{0.95}    & 0.84          \\
\textbf{$\lambda = 8$} & \textbf{0.99}        & 0.97               & \textbf{0.91}                                                     & \textbf{0.89}      & \textbf{0.89}    & \textbf{0.58}                                                          & \textbf{0.68}                                                            & \textbf{0.83}                                                            & 0.79                                                                      & 0.55                                                                     & 0.75                                                                      & 0.93             & \textbf{0.86} \\
\textbf{$\lambda = 16$} & 0.96                 & 0.96               & 0.84                                                              & 0.83               & 0.77             & 0.52                                                                   & 0.64                                                                     & 0.81                                                                     & 0.80                                                                       & 0.53                                                                     & 0.59                                                                      & 0.91             & 0.82          \\ \bottomrule
\end{tabular}}
\end{table*}
\begin{table*}[h]
\caption{Ablation study of Gabor filtering-based anti-aliasing mechanism and fractal dimension-based self-attention distilling}
\label{tab:ablation}
\resizebox{\textwidth}{!}{
\begin{tabular}{@{}l|l|lllllllllllll@{}}
\toprule
\textbf{Ablation Study}                                                                                                       & \textbf{Methods}                      & \textbf{approaching} & \textbf{departing} & \textbf{\begin{tabular}[c]{@{}l@{}}hand\_\\ shaking\end{tabular}} & \textbf{high five} & \textbf{hugging} & \textbf{\begin{tabular}[c]{@{}l@{}}kicking\_\\ left\_leg\end{tabular}} & \textbf{\begin{tabular}[c]{@{}l@{}}kicking \_\\ right\_leg\end{tabular}} & \textbf{\begin{tabular}[c]{@{}l@{}}pointing\_\\ left\_hand\end{tabular}} & \textbf{\begin{tabular}[c]{@{}l@{}}pointing\_ \\ right\_hand\end{tabular}} & \textbf{\begin{tabular}[c]{@{}l@{}}punching\_\\ left\_hand\end{tabular}} & \textbf{\begin{tabular}[c]{@{}l@{}}punching\_ \\ right\_hand\end{tabular}} & \textbf{pushing} & \textbf{OA}   \\ \midrule
\multirow{4}{*}{\textbf{\begin{tabular}[c]{@{}l@{}}Gabor filtering\\ based\\ anti-aliasing\\ mechanism\end{tabular}}}       & \textbf{Baseline1}                    & 0.96                 & 0.96               & 0.79                                                              & 0.85               & 0.69             & 0.55                                                                   & 0.65                                                                     & 0.66                                                                     & 0.65                                                                       & 0.54                                                                     & 0.58                                                                       & \textbf{0.93}    & 0.78          \\
                                                                                                                              & \textbf{Baseline1+Anti-alasing}       & 0.97                 & \textbf{0.98}      & 0.83                                                              & 0.91               & \textbf{0.91}    & \textbf{0.61}                                                          & 0.63                                                                     & 0.74                                                                     & 0.78                                                                       & 0.46                                                                     & 0.69                                                                       & \textbf{0.93}    & 0.84          \\
                                                                                                                              & \textbf{Baseline1+Gabor}              & \textbf{1.00}        & 0.95               & 0.83                                                              & \textbf{0.92}      & 0.67             & 0.53                                                                   & 0.63                                                                     & \textbf{0.85}                                                            & \textbf{0.92}                                                              & 0.42                                                                     & 0.69                                                                       & 0.90             & 0.85          \\
                                                                                                                              & \textbf{GANet}                        & 0.99                 & 0.97               & \textbf{0.91}                                                     & 0.89               & 0.89             & 0.58                                                                   & \textbf{0.68}                                                            & 0.83                                                                     & 0.79                                                                       & \textbf{0.55}                                                            & \textbf{0.75}                                                              & \textbf{0.93}    & \textbf{0.86} \\ \midrule
\multirow{4}{*}{\textbf{\begin{tabular}[c]{@{}l@{}}fractal dimension\\ based \\ self-attention\\ distilling\end{tabular}}}    & \textbf{Baseline2}                & 0.91                 & \textbf{0.98}      & 0.84                                                              & 0.85               & 0.74             & 0.57                                                                   & 0.54                                                                     & 0.71                                                                     & 0.67                                                                       & 0.51                                                                     & 0.62                                                                       & 0.90             & 0.79          \\
                                                                                                                              & \textbf{Baseline2+FrequencyAttention} & 0.97                 & 0.94               & 0.75                                                              & 0.80               & 0.71             & 0.57                                                                   & 0.56                                                                     & 0.84                                                                     & 0.81                                                                       & 0.45                                                                     & 0.61                                                                       & 0.90             & 0.82          \\
                                                                                                                              & \textbf{Baseline2+TemporalAttention}  & 0.79                 & 1.00               & 0.95                                                              & \textbf{0.90}      & 0.86             & 0.50                                                                   & \textbf{0.91}                                                            & \textbf{0.89}                                                            & \textbf{0.91}                                                              & \textbf{0.64}                                                            & 0.50                                                                       & 0.62             & 0.84          \\
                                                                                                                              & \textbf{GANet}                        & \textbf{0.99}        & 0.97               & \textbf{0.91}                                                     & 0.89               & \textbf{0.89}    & \textbf{0.58}                                                          & 0.68                                                                     & 0.83                                                                     & 0.79                                                                       & 0.55                                                                     & \textbf{0.75}                                                              & \textbf{0.93}    & \textbf{0.86} \\ \bottomrule
\end{tabular}}
\end{table*}

\subsection{Comparison with state-of-the-art methods}
\subsubsection{Quantitative Results}
 We compare GraSens with several state-of-the-art approaches on all four datasets, namely WVAR, WAR, HHI, and CSLOS. Apart from SVM and WNN, we used the reported accuracy of their original paper unless otherwise stated for comparison.

Table~\ref{tab:WVAR} illustrates the classification accuracy of the dataset WVAR. GraSens surpassed all other methods in most of the actions with an OA of $95\%$, which is slightly higher than these of SVM and WNN $1\%$. The reason behind this may be due to the fact that the dataset WVAR is relatively too small to reflect the advantages of GraSens. In addition, it can be observed that some action classes (i.e. push, phone talk, and drink) of GraSens obtained a slightly lower accuracy than WNN. The possible reason for this can be that all are simple activities whose changes in waveform characteristics over time were similar. Compared with WNN, GraSens has  fewer advantages in this case.

Table~\ref{tab:WAR} shows the results on the dataset WAR. GraSens outperforms all the baselines with a large margin of $5\%$ than LSTM and $1\%$ than our baseline WNN. Notably, WNN has the same network structure as GraSens. This confirms the effectiveness of the design of our network. Compared with the results of RF, HMM, and SVM, the results of GraSens had obvious improvements in all the six activities. This reason behind this is due to the fact that GraSens can extract more robust and shift-invariant features than machine learning methods. Compared to WNN and LSTM, GraSens achieved the best performance on fall, sit-down, and stand-up, which means that GraSens can capture the characteristics of rapidly changing motion in time and space. These results demonstrated that the GraSens is able to explore the frequency and temporal continuity inside WiFi signals to enhance the quality of output features scattered in different subcarriers. As for lie-dow, GraSens obtained slightly lower but similar performance with $1\%$ than LSTM. The reason is due to that the signals change fast at the beginning but keep similar after in space. With regard to the action walk which behaved similarly in time and space, the accuracy of GraSens was $8\%$ lower than LSTM. The possible reason is that the spectrum of the signals behaves similarly in time. The results indicated that GraSens is good at sophisticated action recognition but slightly poor at simple actions.

Table~\ref{tab:HHI} shows the classification accuracy of the dataset HHI. GraSens obtains the most satisfying results by obvious margins and surpassed the original method E2EDLF. GraSens outperforms the WNN with $7\%$ which confirms the effectiveness of fractal dimension-based self-attention as well as Gabor filtering-based anti-aliasing. Specifically, for the actions of approaching and departing, all of these methods achieved satisfied accuracy over $90\%$. On the basis of the results of hand-shaking, high five, hugging, and pushing, the proposed GraSens outperformed other algorithms. However, the evaluation of GraSens on kicking, pointing and punching lacked effectiveness. The possible reason is that these actions were single limb linear movements and last shortly in time series sequences thus the input CSI samples contained an amount of the noises included in the ambient environment. GraSens augmented the characteristics of WiFi signals and was inevitably affected by these noises. Overall, the performance of GraSens was moderate, but it was still more convenient to realize action recognition with no requirements for the sophisticated preprocessing than the state-of-art E2EDLF, especially on complex actions in the temporal and frequency domains.

\subsubsection{Qualitative Results}
We also show the effectiveness of WiFi and Video data on WVAR. Fig.~\ref{fig:csi_pre}(b) and (c) illustrate that CSI signals are not affected by the occlusion and exhibit similar patterns in the same actions. 

 Skeleton visualization is further to show the effectiveness of WVAR. Inspired by the work~\cite{zhao2018through}, the skeletons derived from Alphapose~\cite{fang2017rmpe} are used to train the GraSens in LOS conditions. On the basis of the skeletons, the trained  GraSens can further generate skeletons in non-line-of-light scenes. Skeleton visualization is further to show the effectiveness of WVAR. As seen in Fig.~\ref{fig:skeleton}(a)-(d), in the scene without any occlusions, our GraSens yielded robust skeletons in good agreement with the truth images which were close to these of Alphapose. In partially covered situations, GraSens provided the most convincing skeleton results such as seating in Fig.~\ref{fig:skeleton}(d) compared to Alphapose in Fig.~\ref{fig:skeleton}(h), with the skeleton boundary being visually close to the raw truth image. This clearly demonstrates that our CSI data on WVAR has a good efficiency in these scenarios.

\subsection{Ablation Study}

In this subsection, we have implemented the experiments to reveal how the different number of GraSens blocks influence the classification accuracy. In addition, we also conducted additional experiments on GraSens with ablation consideration. In this study, we use HHI as the benchmark to test the additional effects of the different number of GraSens blocks as well as self-attention and anti-aliasing mechanisms.

\subsubsection{The performance of number of GraSens blocks}
The number of stacked blocks $\lambda$ has a trade-off between the accuracy and efficiency of the proposed GraSens method. To further verify the influence of the number of stacked blocks on performance, we have added an experiment as illustrated in Table V. As shown in Table V, the GraSens achieves the better performance with a growth of $2\%$ when $\lambda =8$ compared with when $\lambda =4$. In contrast, when we add the number of blocks to $\lambda =16$, the classification accuracy decreases by $2\%$. It is noted that the 16 GraSens blocks network architecture is over-fitting for the training data and generalizes poorly on new testing data. As a result, the classification accuracy decreases on the contrary. According to the results, we choose $\lambda =8$ as the number of blocks used in our experiments empirically.

\subsubsection{The performance of Gabor filtering-based anti-aliasing mechanism}
In this study, we testify to the potential accuracy of our Gabor filtering, anti-aliasing, and Gabor filtering-based anti-aliasing in acquiring “generative” results illustrated in Table~\ref{tab:ablation}. Firstly, WNN with the fractal dimension-based self-attention is set as the main pipeline 'baseline1'. For the second, we replace the pooling with an anti-aliasing operation. For the third, the Gabor filtering replaces the first layer of baseline as the Gabor convolution layer. Surprisingly, both anti-aliasing operation and Gabor filtering largely improve the classification accuracy by $8\%$ and $9\%$, respectively. In addition, the fusion of two operations continues to enhance the performance by $9\%$.This confirms both the correlation between Gabor filtering and anti-aliasing operation and the importance of the fusion of each other. Thereafter, Gabor filtering-based anti-aliasing further improves the performance, widening the gap with the existing methods.

\subsubsection{The performance of fractal dimension-based self-attention distilling}
In the overall results Table~\ref{tab:ablation}, we distill frequency and temporal attention separately for self-attention. Firstly, WNN with Gabor filtering-based anti-aliasing is used as the 'baseline2'. Firstly, we add the baseline2 with fractal dimension frequency attention only. As for the second, we add the baseline2 with fractal dimension temporal attention. The fractal dimension-based self-attention determines how the network distributes the contribution of the features. We notice that both the frequency attention and the temporal attention contribute to the improvements of accuracy by 3$\%$ and 4$\%$. The integration of both can further refine the accuracy by 7$\%$.

\section{Conclusion}
In this paper, we identified the inherent limitation of the WiFi signal-based convolution neural networks, with observations that the efficacy of WiFi signals is prone to be influenced by the change in the ambient environment and varies over different sub-carriers. Thereafter, based on their characteristics, we proposed to formulate reliable and robust temporal and frequency shift-invariant representations. We first designed the Gabor filtering based on anti-aliasing to obtain the shift-invariant feature information of actions with the strong auxiliary function. Furthermore, fractal dimension-based frequency and temporal self-attention are proposed to focus on the dominant features scattered in different subcarriers. In addition, we collected synchronous video and WiFi datasets WVAR to simulate the complex visual conditions like the occlusions scenarios. The ablation study verified that both our Gabor filtering-based anti-aliasing and fractal dimension-based frequency and temporal self-attention are beneficial for the improvement of classification accuracy. Through the experiments on the four most popular datasets, our GraSens achieved a new state-of-the-art with a large margin. We believe it would be a promising future direction to adopt the Gabor filtering-based anti-aliasing and fractal dimension-based attention to the HAR or other related tasks.

\bibliographystyle{IEEEtran}
\bibliography{IEEEtran}

\end{document}